\documentclass[letterpaper, 10 pt, conference]{ieeeconf}  

\IEEEoverridecommandlockouts                              

\usepackage{graphicx} 
\usepackage{amsmath} 
\usepackage{amssymb}  

\usepackage{xspace}

\makeatletter
\DeclareRobustCommand\onedot{\futurelet\@let@token\@onedot}
\def\@onedot{\ifx\@let@token.\else.\null\fi\xspace}

\def\eg{\emph{e.g}\onedot}

\def\etc{\emph{etc}\onedot}

\makeatother

\title{\LARGE \bf CPGNet: Cascade Point-Grid Fusion Network for Real-Time LiDAR Semantic Segmentation}

\author{Xiaoyan Li$^{*,+}$, Gang Zhang$^{*,\times}$, Hongyu Pan$^{\times}$, Zhenhua Wang$^{\times}$
\thanks{$^{+}$University of Chinese Academy of Sciences, Beijing 100049, China.
{\tt\small xiaoyan.li@vipl.ict.ac.cn}}%
\thanks{$^{\times}$Damo Academy, Alibaba Group.
\newline {\tt\small zhanggang11021136@gmail.com}
\newline {\tt\small hongyu.pan@alibaba-inc.com}
\newline {\tt\small zhwang.me@gmail.com}}%
\thanks{* equal contribution}
}

\begin{document}

\maketitle
\thispagestyle{empty}
\pagestyle{empty}

\begin{abstract}
LiDAR semantic segmentation essential for advanced autonomous driving is required to be accurate,
fast, and easy-deployed on mobile platforms. Previous point-based or sparse voxel-based methods are far away from
real-time applications since time-consuming neighbor searching or sparse 3D convolution are employed. Recent 2D projection-based methods,
including range view and multi-view fusion, can run in real
time, but suffer from lower accuracy due to information loss during the 2D projection. Besides, to improve the performance, previous methods usually adopt test time augmentation (TTA), which further slows down the inference process. To achieve a better speed-accuracy trade-off, we propose Cascade Point-Grid Fusion Network (CPGNet), which ensures both effectiveness and efficiency mainly by the following two techniques: 1) the novel Point-Grid (PG) fusion block extracts semantic features mainly on the 2D projected grid for efficiency, while summarizes both 2D and 3D features on 3D point for minimal information loss; 2) the proposed transformation consistency loss narrows the gap between the single-time model inference and TTA. The experiments on the SemanticKITTI and nuScenes benchmarks demonstrate that the CPGNet without ensemble models or TTA is comparable with the state-of-the-art RPVNet, while it runs $4.7$ times faster.
\end{abstract}

\section{INTRODUCTION}
Light Detection and Ranging (LiDAR) sensors are widely used in autonomous driving and robotics. The 3D point cloud data they captured provide rich information about the surrounding scene. LiDAR semantic segmentation assigns semantic labels for these 3D point clouds, such as car, pedestrian, cyclist, road, building, and thus directly bears on driving accuracy and safety. In the last few years, various deep learning models have been proposed to process LiDAR 3D point cloud, but these methods can not guarantee accuracy and speed simultaneously, especially on mobile platforms (\eg cars and robots).

\begin{figure}[t]
\centering
\includegraphics[width=0.9\linewidth]{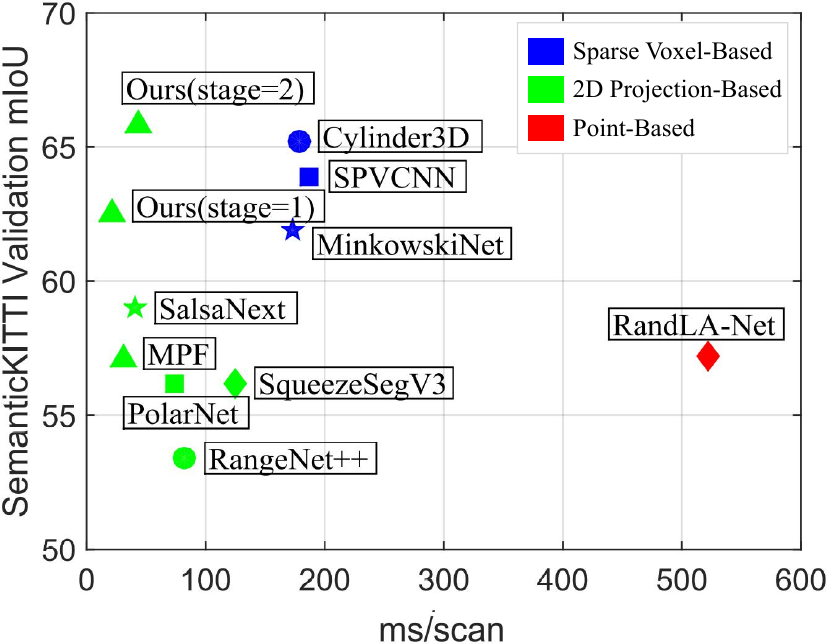}
\caption{Accuracy (mIoU) vs. runtime on the SemanticKITTI validation set. For fair comparison, all methods are trained based on the official code and evaluated without ensemble models or TTA. The experiments are conducted with PyTorch FP32 on NVIDIA RTX 2080Ti GPU.}
\label{Fig_iou_vs_speed}
\end{figure}

Previous methods on LiDAR 3D point cloud can be grouped into three categories: point-based methods, sparse voxel-based methods, and 2D projection-based methods.

The point-based methods include PointNet \cite{qi2017pointnet}, PointNet++ \cite{qi2017pointnetplus}, PointCNN \cite{li2018pointcnn}, RandLA-Net \cite{hu2020randla} and \etc They are directly applied on the raw unordered 3D point cloud without any information loss. However, these methods usually adopt time-consuming operations, namely farthest point sampling (FPS) for uniformly down-sampling, and k-nearest neighbor (kNN) or ball query for local neighbor searching.

Since LiDAR 3D points are hugely sparse, the sparse voxel-based methods quantize the 3D points to voxels and then apply 3D convolution operation only on these sparse voxels. Although these methods have inevitable information loss due to quantization, they achieve the state-of-the-art performance. However, these methods are computationally expensive, and can not run in real-time.

The 2D projection-based methods apply mature 2D CNN on the 2D grid feature map projected by the 3D point cloud. Inspired by Fully Convolution Network (FCN) \cite{long2015fully} and its variants \cite{noh2015learning, zhao2017pyramid, chen2017deeplab, chen2017rethinking, chen2018encoder} that dominate 2D image semantic segmentation, these 2D projection-based methods, usually considering bird’s-eye view \cite{lang2019pointpillars} or range view \cite{wang2018pointseg}, can be easily designed and deployed on some efficient deep learning inference frameworks (\eg TensorRT \cite{vanholder2016efficient}). They can achieve almost 15\,ms per LiDAR scan. However, these methods have lower accuracy due to serious 2D projection information loss. They still perform worse, even if the following RangeNet++ \cite{milioto2019rangenetplus} attempts to use kNN as post-process, and MPF \cite{alnaggar2021multi} combines both bird’s-eye view and range view.

To this end, we propose the CPGNet for real-time accurate LiDAR semantic segmentation. The proposed PG fusion block in CPGNet first projects and extracts semantic features on the 2D grids of both bird’s-eye view and range view, and then transmits and fuses these features onto the 3D point. As can be seen, the PG fusion block combines the advantages of both point-based methods with complete information and 2D projection-based methods with fast speed. The CPGNet applies PG fusion block repeatedly to further enhance point features. Besides, inspired by test time augmentation (TTA), the transformation consistency loss is proposed to ensure agreements between the results of original and augmented point clouds. Finally, we 
compare CPGNet with the open-source methods, as shown in Fig. \ref{Fig_iou_vs_speed}, and CPGNet achieves the best mIoU (65.9) on SemanticKITTI \cite{behley2019semantickitti} validation set, while it runs 43\,ms with PyTorch FP32 on NVIDIA RTX 2080Ti GPU. The contributions can be listed as follows:

\begin{itemize}
\item We present an accurate, fast, and easy-deployed CPGNet
for LiDAR semantic segmentation. It fuses point, bird’s-eye view and range view features in a cascade framework.

\item We propose transformation consistency loss inspired by test time augmentation (TTA), and enable higher performance  with only single-time inference.

\item The proposed CPGNet achieves the best speed-accuracy trade-off on SemanticKITTI and nuScenes benchmarks.
\end{itemize}

\section{RELATED WORK}
Unlike 2D images with dense grid structures, point clouds are unordered, sparse, and unstructured, which make it difficult to apply deep learning operations (\eg convolution). Previous methods attempt to solve this problem in three ways.

\subsection{Point-Based Methods}
Point-based methods directly operate on the raw points. The PointNet \cite{qi2017pointnet} applies shared Multi-Layer Perceptron (MLP) on each point and max pooling among the whole points to acquire point-level features for further segmentation task. However, PointNet performs worse on the complicated scene for the lack of local context extraction. The following work \cite{qi2017pointnetplus, li2018pointcnn} propose ball query and $\chi$-Conv to mimic 2D convolution, and achieve great results on the indoor scene. However, they cannot be applied on the LiDAR point cloud due to computation and memory cost. To accelerate the network inference, RandLA-Net \cite{hu2020randla} adopts random sampling and local feature aggregation, but it suffers from lower accuracy due to random sampling. KPConv \cite{thomas2019kpconv} proposes a novel spatial kernel-based
point convolution to extract local structure, and KPRNet \cite{kochanov2020kprnet} combines KPConv and ResNext \cite{xie2017aggregated} to achieve the best results of point-based methods. Although point-based methods are directly applied on the raw points without dropping the information out, it’s less studied in autonomous driving due to its inefficient local structure extraction.

\subsection{Sparse Voxel-Based Methods}
Sparse voxel-based methods quantize the 3D points into sparse voxels, and then apply 3D convolution operation only on those non-empty voxels to reduce computation and memory cost. Minkowski CNN \cite{choy20194d} is the first efficient sparse voxel framework, and it surpasses all point-based methods in both accuracy and speed. A possible reason is that sparse voxel is structured, which is convenient for convolution operation. SPVNAS \cite{tang2020searching} introduces neural architecture search (NAS) into \cite{choy20194d}, and achieves better results with lower computation cost. Recently, variants \cite{zhu2021cylindrical, cheng20212, xu2021rpvnet} of sparse voxel-based methods are proposed. Cylinder3D \cite{zhu2021cylindrical} quantizes the 3D points in the cylindrical coordinate system, and proves its efficiency. AF2S3Net \cite{cheng20212} proposes the attentive feature fusion module (AF2M) and adaptive feature selection module (AFSM) to efficiently extract local and global structures simultaneously. RPVNet \cite{xu2021rpvnet} fuses range view, point, and sparse voxel features in a single framework to alleviate quantization error, and achieves the best results on the SemanticKITTI and nuScenes benchmarks. Although these methods dominate the LiDAR semantic segmentation benchmarks, they have difficulty in deployment and cannot run in real-time on mobile platforms.

\subsection{2D Projection-Based Methods}
Recently 2D projection-based methods attract more attentions because they are fast and easy-deployed. These methods utilize 2D FCN by projecting 3D points onto the 2D grid and primarily include range view and multi-view fusion. Range view projects 3D points onto the 2D spherical grid and has many variants. RangeNet++ \cite{milioto2019rangenetplus} proposes an accelerated kNN for post-process, which is an indispensable module in the following range-based methods. SqueezeSegV3  \cite{xu2020squeezesegv3} proves the superiority of the spatially adaptive convolution. SalsaNext \cite{cortinhal2020salsanext} designs a novel encoder-decoder network based on SalsaNet \cite{aksoy2020salsanet}, and adopts Lov{\'a}sz-Softmax loss \cite{berman2018lovasz} that can directly optimize the mean Intersection over Union (mIoU) metric. More recently, Lite-HDSeg \cite{razani2021lite} proposes harmonic dense convolutions and achieves the best results among range-based methods. Since the single view or 2D grid has inevitable 2D projection information loss, the following multi-view fusion projects the 3D points to two or more different types of 2D grid. MPF \cite{alnaggar2021multi} and AMVNet \cite{liong2020amvnet} both combine bird’s-eye view and range view. Different from the proposed method, they conduct semantic segmentation independently on each view and fuse the segmentation results from the two views by a late-fusion module.

\section{PROPOSED METHOD}

\begin{figure*}[t]
\centering
\includegraphics[scale=0.75]{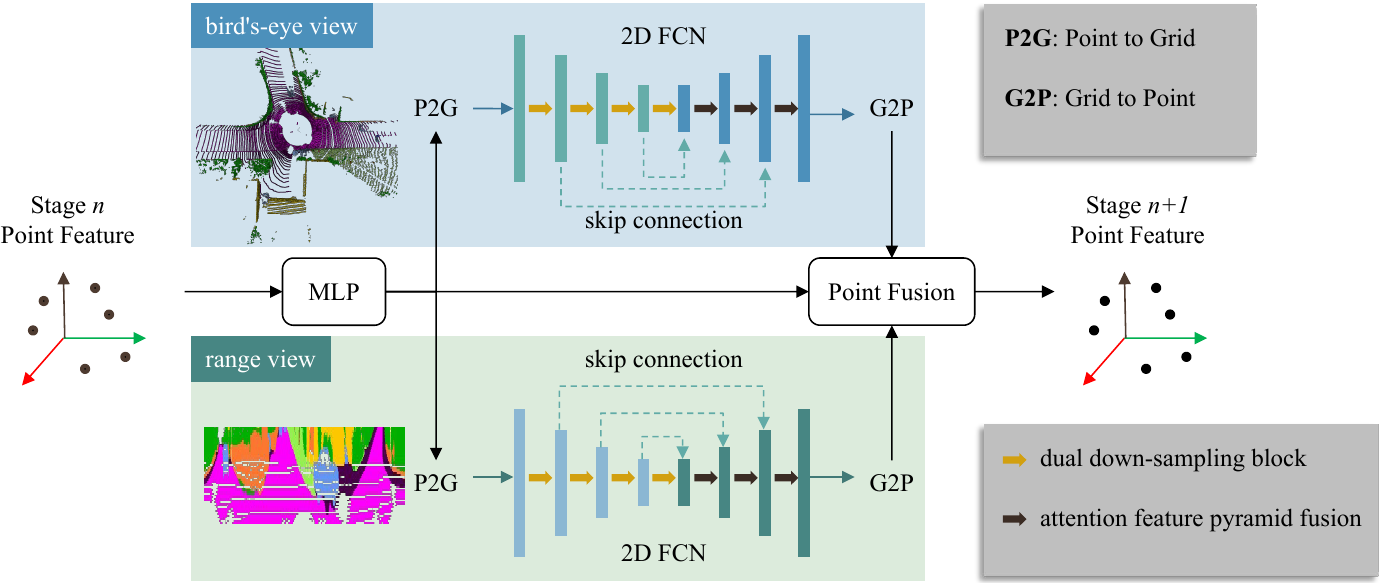}
\caption{The Point-Grid (PG) fusion block. It takes the point features from the last PG fusion block as input and undergoes the point, bird’s-eye view and range view branches, respectively. The output point features are acquired by fusing features from the three branches.}
\label{Fig_cpgnet}
\end{figure*}

For accurate and fast LiDAR semantic segmentation, it needs to not only extract semantic features in an efficient way but also keep complete point cloud information. Therefore, we propose CPGNet which progressively extracts point features by the Point-Grid (PG) fusion block. As shown in Fig. \ref{Fig_cpgnet}, the PG fusion block consists of four steps: 1) the Point to Grid ({\bf P2G}) operation projects the input point features onto bird’s-eye view and range view feature maps respectively; 2) the 2D FCN is applied on the 2D feature maps to extract semantic features efficiently; 3) the Grid to Point ({\bf G2P}) operation transmits the 2D grid features onto the 3D point; 4) the Point Fusion fuses features from the 3D point, bird’s-eye view and range view branches to ensure complete point cloud information. As can be seen, CPGNet combines the advantages of both point-based and 2D projection-based methods. The components of CPGNet, except {\bf P2G} and {\bf G2P}, can be directly deployed on TensorRT. The {\bf P2G} and {\bf G2P} operations can be implemented via efficient CUDA code. Each PG fusion block of CPGNet shares the same network architecture, but does not share parameters. The details are described in the subsections below.

\subsection{Point to Grid}
\begin{figure}[t]
\centering
\includegraphics[scale=0.55]{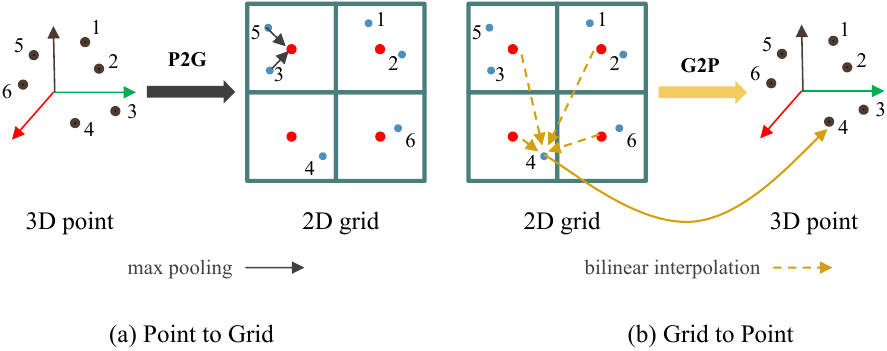}
\caption{The diagram of Point to Grid (a) and Grid to Point (b) operations.}
\label{Fig_p2g}
\end{figure}

The Point to Grid ({\bf P2G}) operation aims to transform 3D point features to 2D grid feature maps. As shown in Fig. \ref{Fig_p2g}a, it first projects the $k^{th}$ 3D point $\boldsymbol{p}_k^{3D}=(x_k,y_k,z_k)$ onto 2D grid to acquire the corresponding 2D coordinates $\boldsymbol{p}_k^{2D}=(u_k,v_k)$. The set $\mathcal{R}_{h,w}$ contains the indices of points that fall in the same 2D grid $(h, w)$, namely $\mathcal{R}_{h,w}=\{k|\lfloor u_k \rfloor = h; \lfloor v_k \rfloor = w \}$. Then the features $\boldsymbol{\mathcal{F}}_k^{3D}$ of point in $\mathcal{R}_{h,w}$ are gathered through max pooling to form the corresponding 2D grid features $\boldsymbol{\mathcal{G}}^{2D}_{h,w}$. The formula is as follows:
\begin{equation}
\boldsymbol{\mathcal{G}}^{2D}_{h,w,c} = \max \limits_{k \in \mathcal{R}_{h,w}} \boldsymbol{\mathcal{F}}^{3D}_{k,c}.
\label{equat_maxpool}
\end{equation}
There may be more than one points falling in a 2D grid. To avoid parallel conflicts, while processing the same 2D grid, it utilizes CUDA atomicMax function.

Both bird’s-eye view and range view are used in the proposed method. Bird’s-eye view omits $z$ dimension, and range view discards $r$ dimension. Therefore, the two views are complementary to alleviate 2D projection information loss. Actually, the two views use a similar {\bf P2G} operation. They just differs in the way of 2D projection. For bird’s-eye view, it projects the 3D point onto $x-y$ plane that is discretized using a rectangular 2D grid
$(x_{min}, y_{min}, x_{max}, y_{max})$ with the predefined width $W_{bev}$ and height $H_{bev}$, as summarized in the following equation,
\begin{equation}
\begin{pmatrix}
u_k \\ \\ v_k
\end{pmatrix}=
\begin{pmatrix}
\frac{x_k - x_{min}}{x_{max} - x_{min}} \times W_{bev} \\ \\
\frac{y_k - y_{min}}{y_{max} - y_{min}} \times H_{bev}
\end{pmatrix}.
\end{equation}
For range view, the 3D point is mapped from the 3D cartesian space $\boldsymbol{p}_k^{3D} = (x_k,y_k,z_k)$ to the spherical space $\boldsymbol{p}_k^{sph}=(r_k,\theta_k,\phi_k)$ by applying
\begin{equation}
\begin{pmatrix}
r_k \\ \\ \theta_k \\ \\ \phi_k
\end{pmatrix}=
\begin{pmatrix}
\sqrt{{x_k}^2 + {y_k}^2 + {z_k}^2} \\ \\
\arcsin{(\frac{z_k}{\sqrt{{x_k}^2 + {y_k}^2 + {z_k}^2}})} \\ \\
\arctan{(y_k, x_k)}
\end{pmatrix},
\end{equation}
where $r_k$, $\theta_k$, $\phi_k$ denote the distance, zenith and azimuth angle respectively. Subsequently, range view grid with the predefined width $W_{rv}$ and height $H_{rv}$ is acquired by discretizing $\theta_k$ and $\phi_k$ but ignoring $r_k$,
\begin{equation}
\begin{pmatrix}
u_k \\ \\ v_k
\end{pmatrix}=
\begin{pmatrix}
\frac{1}{2}[1 - \phi_k\pi^{-1}]W_{rv} \\ \\
[1-(\theta_k+f_{up})f^{-1}]H_{rv}
\end{pmatrix},
\end{equation}
where $f=f_{up} + f_{down}$ is the LiDAR vertical field-of-view.

\subsection{2D FCN}
\begin{figure}[t]
\centering
\includegraphics[scale=0.55]{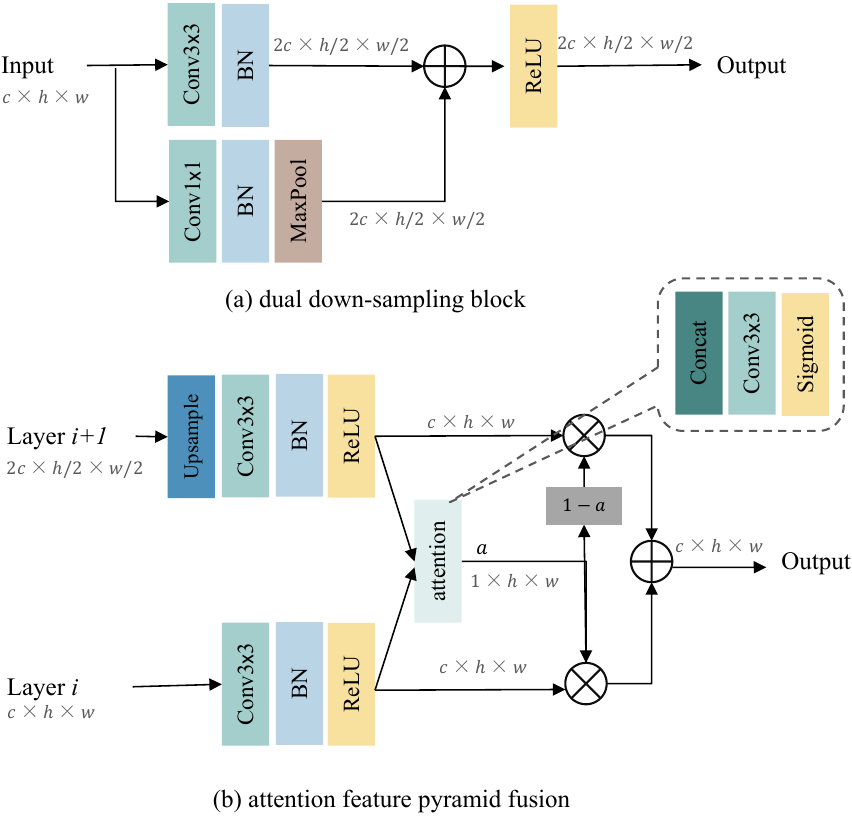}
\caption{(a) dual down-sampling block. (b) attention feature pyramid fusion. Note that $c$, $h$ and $w$ denotes channels, height and width of the 2D feature map, respectively.}
\label{Fig_fcn}
\end{figure}

The 2D FCNs with the encoder and decoder architectures, are applied on the bird’s-eye view and range view feature maps respectively to extract semantic features. They occupy more than $90\%$ computation costs of the CPGNet. Therefore, the encoder network utilizes ResNet \cite{he2016deep} with only 9 layers as a lightweight backbone, and has 128 maximum number of channels. To keep information during down-sampling, we propose the dual down-sampling block that uses 2D Convolution and 2D MaxPool for down-sampling in parallel, as shown in Fig. \ref{Fig_fcn}a. In the experiment, it demonstrates that the dual down-sampling block performs better with negligible latency.

The previous decoder architectures \cite{chen2018encoder, cortinhal2020salsanext} usually adopt feature pyramid fusion to fuse high-level and low-level feature maps. High-level feature maps contain more semantic information, while low-level feature maps show more details. For semantic segmentation, some parts (\eg road, building) need high-level semantic features, and some parts (\eg pedestrian, object boundary) require detailed features. Instead of using simple feature maps concatenation, we propose attention feature pyramid fusion to automatically select features from different levels, as shown in Fig. \ref{Fig_fcn}b.

\subsection{Grid to Point}
On the contrary to the Point to Grid ({\bf P2G}) operation, the Grid to Point ({\bf G2P}) transfers features from 2D grid to 3D point, according to the corresponding 2D coordinates $\boldsymbol{p}^{2D}_k=(u_k,v_k)$. As shown in Fig. \ref{Fig_p2g}b, it applies bilinear interpolation within the four neighbor grids. The formula is as follows:
\begin{align}
\boldsymbol{\mathcal{F}}^{3D}_{k,c} = \sum_{i=0}^{1} \sum_{j=0}^{1} w_{i,j,k} \boldsymbol{\mathcal{G}}^{2D}_{\lfloor u_k \rfloor+i,\lfloor v_k \rfloor + j,c},
\end{align}
where $w_{i,j,k} = (1-|u_k - (\lfloor u_k \rfloor + i)|)(1-|v_k - (\lfloor v_k \rfloor + j)|)$ denotes the bilinear interpolation weight. Note that the neighbor grids beyond the 2D grid range are regarded as all zeros. As can be seen, each point and each feature channel are calculated independently, which is much suitable for the CUDA parallel computing.

\subsection{Point Fusion}
The Point Fusion module is responsible for fusing the features from the point, bird’s-eye view and range view. For efficiency, it only adopts feature concatenation and two MLP layers. Different from MPF \cite{alnaggar2021multi} and AMVNet \cite{liong2020amvnet}, the Point Fusion does not conduct in post-process but serves as a mid-fusion module that is an important part of the proposed end-to-end CPGNet. The end-to-end framework has two advantages: 1) it is convenient to be deployed with rare post-process; 2) it can narrow the gap between the training and evaluation phases. The experiment section proves its superiority.

Though the features of the point that is projected out of the 2D grid range are regarded as all zeros in a certain view, it can pass information from the other view. For instance, the point that is beyond the range of bird’s-eye view, but in the range of range view, has meaningful features from range view. In the experiment, we analyze the point distributions on both views and find that almost all points fall in the range of at least one view.

\subsection{Loss Function}
The segmentation predictions are acquired by applying a Fully Connection (FC) layer to the output features of the PG fusion block. The LiDAR semantic segmentation dataset (\eg SemanticKITTI, nuScenes) has highly unbalanced categories. For instance, the proportions of $road$, $sidewalk$ and $building$ are hundreds times than that of $person$ and $motorcyclist$. To this end, we adopt weighted cross entropy (WCE) loss to manually emphasize rare categories. The WCE loss can be formulated as
\begin{align}
\alpha_c =& \frac{1}{F_c+\epsilon} \nonumber \\
\mathcal{L}_{wce} =& -\sum_{c=1}^{C}\alpha_c y_c \log(\hat{y}_c),
\end{align}
where $y_c$ defines the ground-truth label, $\hat{y}_c$ is the predicted probability, $F_c$ is the frequency, and $\alpha_c$ is the weight of the $c^{th}$ class. $C$ is the category number of the dataset. In the experiment, $\epsilon$ is set as $0.001$. We also adopt Lov{\'a}sz-Softmax loss \cite{berman2018lovasz} that can optimize the mean Intersection over Union (mIoU) metric as the second loss term $\mathcal{L}_{ls}$. As shown in \cite{aksoy2020salsanet, razani2021lite}, it really improves the mIoU metric of segmentation task. More details can be seen in \cite{berman2018lovasz}.

Previous methods \cite{tang2020searching, zhu2021cylindrical} adopt test time augmentation (TTA) to improve the performance, which needs model inference for multiple times. For both effectiveness and efficiency, we propose transformation consistency loss $\mathcal{L}_{tc}$ to reduce the differences between raw points and augmented points. The formula is as follows:
\begin{equation}
\mathcal{L}_{tc}=\sum_{c=1}^{C}|\hat{y}_c^{raw} - \hat{y}_c^{aug}|,
\end{equation}
where $\hat{y}_c^{raw}$, $\hat{y}_c^{aug}$ are the predicted probability respectively from raw points and augmented points. The total loss $\mathcal{L}_{total}$ is the sum of the three loss terms and is defined as
\begin{equation}
\mathcal{L}_{total}=\mathcal{L}_{wce} + 2\mathcal{L}_{ls} + \mathcal{L}_{tc}.
\end{equation}

\section{EXPERIMENTAL RESULTS}
We evaluate the performance of the proposed CPGNet on SemanticKITTI \cite{behley2019semantickitti} and nuScenes \cite{caesar2020nuscenes} benchmarks.

{\bf SemanticKITTI.} It contains 43,552 $360^\circ$ LiDAR scans from 22 sequences collected in a city of Germany. Equipped with a Velodyne HDL-64E rotating LiDAR with 64 beams vertically, each LiDAR scan has approximately 130k points. The training set (19,130 scans) consists of sequence from 00 to 10 except 08, and the sequence 08 (4,071 scans) is used for validation. The rest sequences (20,351 scans) from 11 to 21 are only provided with LiDAR point clouds, and are used for online leaderboard. This dataset is labeled with 28 classes, but a high-level label set of 19 classes is used for single-scan LiDAR semantic segmentation.

{\bf nuScenes.} It is a newly released dataset for LiDAR semantic segmentation with 1,000 scenes collected from different areas of Boston and Singapore. Each scene is 20s long collected by a Velodyne HDL-32E rotating LiDAR with 32 beams vertically. It splits 28,130 samples for training, 6,019 for validation and 6,008 for testing. It is annotated with 32 classes, and just 16 classes are used for official evaluation after merging some similar classes and removing rare classes.

{\bf Evaluation Metric.} We adopt the most popular metric, mean Intersection over Union (mIoU), to evaluate the proposed CPGNet and its competitors. It can be formulated as
\begin{equation}
mIoU = \frac{1}{C}\sum_{c=1}^{C}\frac{TP_{c}}{TP_{c} + FP_{c} + FN_{c}},
\end{equation}
where $TP_{c}$, $FP_{c}$, $FN_{c}$ are the true positive, false positive and false negative of $c^{th}$ class respectively. $C$ is the total number of classes.

\subsection{Experimental Setup}
{\bf Network Setup.}
As shown in Fig. \ref{Fig_cpgnet}, each PG fusion block of the CPGNet has a similar network architecture but different parameters. In the experiment, we adopt two cascade PG fusion blocks. The numbers of the input point feature channels for these two blocks are 7 and 64, respectively. The input 7 channels of the first block refer to $x$, $y$, $z$, $intensity$, $r$, $\Delta x$, and $\Delta y$, where $\Delta x$ and $\Delta y$ denote the offsets to the corresponding 2D BEV grid center. The first MLP layer of each block outputs 64 feature channels, which the following {\bf P2G} operation transforms to bird’s-eye view and range view feature maps. The two views utilize a similar 2D FCN network with three down-sampling and three up-sampling stages, but the range view does not apply down-sampling along the height dimension. The feature channels of each stage in the 2D FCN are 64, 32, 64, 128, 128, 96, 64, and 64, respectively. Therefore, the inputs of Point Fusion are $64 \times 3$ feature channels from the three branches. The output channels of these two PG fusion blocks are 64 and 96, respectively.

For SemanticKITTI, the bird’s-eye view branch accepts a 2D feature map with the shape $(W_{bev}=600, H_{bev}=600)$ and the range $(x_{min}=-50, y_{min}=-50, x_{max}=50, y_{max}=50)$. And the range view branch sets the input shape as $(W_{rv}=2048, H_{rv}=64)$. The nuScenes adopts the same configuration except $H_{rv}=32$. Based on these hyper parameters, we find that $99.99\%$ of the whole points fall in at least one view on SemanticKITTI, as shown in Table \ref{table_in_range}.

\begin{table}[t]
\caption{The proportions of points in the 2D grid range on SemanticKITTI.}
\label{table_in_range}
\begin{center}
\begin{tabular}{c|c|c|c}
\hline
BEV & RV & BEV and RV & BEV or RV\\
\hline
98.76\% & 98.60\% & 97.36\% & 99.99\%\\
\hline
\end{tabular}
\end{center}
\end{table}

{\bf Training Details.}
All experiments are conducted with PyTorch FP32 on NVIDIA RTX 2080Ti GPU. Our CPGNet is trained from scratch for 48 epochs with a batch size of 16, which takes around 24 hours on 8 GPUs. The optimizer utilizes stochastic gradient descent (SGD) with a weight decay of 0.001, a momentum of 0.9, and an initial learning rate of 0.02, which is decayed by 0.1 every 10 epochs. Besides, the data augmentation strategies include random rotation around $z$ axis, random global scale sampled from $[0.95, 1.05]$, random flipping along $x$ and $y$ axes, random Gaussian noise $\mathcal{N}(0, 0.02)$, and instance CutMix~\cite{xu2021rpvnet}.

\subsection{Results}
As shown in Table \ref{table_sk}, we compare the proposed CPGNet with the state-of-the-arts on SemanticKITTI test set. The methods are grouped as point-based, 2D projection-based and sparse voxel-based methods from top to bottom. We find that CPGNet outperforms all point-based and 2D projection-based methods, and it can be comparable with the first ranking RPVNet \cite{xu2021rpvnet} in most categories except $motorcyclist$. This category has less training samples and is confused with $bicyclist$ and $motorcycle$, which can be solved by LiDAR and image fusion. On the category of $truck$ and $traffic$-$sign$, CPGNet surpasses RPVNet by a large margin.

Besides, CPGNet runs much faster than the top ranking methods, including SPVCNN \cite{tang2020searching}, Cylinder3D \cite{zhu2021cylindrical}, DRINet \cite{ye2021drinet} and RPVNet \cite{xu2021rpvnet}. Note that we also test the speed of CPGNet on NVIDIA Tesla V100 GPU (marked by *) for fair comparison with RPVNet.

\begin{table*}[t]
\caption{Class-wise and mean IoU of the proposed CPGNet and the competitors on SemanticKITTI single scan leaderboard. Note that speed measurements are taken on a single NVIDIA RTX 2080Ti GPU, while $*$ means that it uses NVIDIA Tesla V100 GPU.}
\label{table_sk}
\setlength\tabcolsep{2.8pt}
\begin{center}
\begin{tabular}{l|c|c|ccccccccccccccccccc}
\hline
{\bf Methods} & \rotatebox{90}{\bf mIoU} & \rotatebox{90}{\bf speed (ms)} & \rotatebox{90}{Car} & \rotatebox{90}{Bicycle} & \rotatebox{90}{Motorcycle} & \rotatebox{90}{Truck} & \rotatebox{90}{Other-vehicle} & \rotatebox{90}{Person} & \rotatebox{90}{Bicyclist} &
\rotatebox{90}{Motorcyclist} & \rotatebox{90}{Road} & \rotatebox{90}{Parking} & \rotatebox{90}{Sidewalk} & \rotatebox{90}{Other-ground} & \rotatebox{90}{Building} &
\rotatebox{90}{Fence} & \rotatebox{90}{Vegetation} & \rotatebox{90}{Trunk} & \rotatebox{90}{Terrain} & \rotatebox{90}{Pole} & \rotatebox{90}{Traffic-sign} \\
\hline
PointNet \cite{qi2017pointnet} & 14.6 & - & 46.3 & 1.3 & 0.3 & 0.1 & 0.8 & 0.2 & 0.2 & 0.0 & 61.6 & 15.8 & 35.7 & 1.4 & 41.4 & 12.9 & 31.0 & 4.6 & 17.6 & 2.4 & 3.7 \\
PointNet++ \cite{qi2017pointnetplus} & 20.1 & - & 53.7 & 1.9 & 0.2 & 0.9 & 0.2 & 0.9 & 1.0 & 0.0 & 72.0 & 18.7 & 41.8 & 5.6 & 62.3 & 16.9 & 46.5 & 13.8 & 30.0 & 6.0 & 8.9 \\
RandLA-Net \cite{hu2020randla} & 53.9 & 521.8 & 94.2 & 26.0 & 25.8 & 40.1 & 38.9 & 49.2 & 48.2 & 7.2 & 90.7 & 60.3 & 73.7 & 20.4 & 86.9 & 56.3 & 81.4 & 61.3 & 66.8 & 49.2 & 47.7 \\
KPConv \cite{thomas2019kpconv} & 58.8 & - & 96.0 & 30.2 & 42.5 & 33.4 & 44.3 & 61.5 & 61.6 & 11.8 & 88.8 & 61.3 & 72.7 & 31.6 & 90.5 & 64.2 & 84.8 & 69.2 & 69.1 & 56.4 & 47.4 \\
\hline
RangeNet++ \cite{milioto2019rangenetplus} & 52.2 & 82.3 & 91.4 & 25.7 & 34.4 & 25.7 & 23.0 & 38.3 & 38.8 & 4.8 & 91.8 & 65.0 & 75.2 & 27.8 & 87.4 & 58.6 & 80.5 & 55.1 & 64.6 & 47.9 & 55.9 \\
SqueezeSegv3 \cite{xu2020squeezesegv3} & 55.9 & 124.3 & 92.5 & 38.7 & 36.5 & 29.6 & 33.0 & 45.6 & 46.2 & 20.1 & 91.7 & 63.4 & 74.8 & 26.4 & 89.0 & 59.4 & 82.0 & 58.7 & 65.4 & 49.6 & 58.9 \\
SalsaNext \cite{cortinhal2020salsanext} & 59.5 & 40.7 & 91.9 & 48.3 & 38.6 & 38.9 & 31.9 & 60.2 & 59.0 & 19.4 & 91.7 & 63.7 & 75.8 & 29.1 & 90.2 & 64.2 & 81.8 & 63.6 & 66.5 & 54.3 & 62.1 \\
Lite-HDSeg \cite{razani2021lite} & 63.8 & - & 92.3 & 40.0 & 55.4 & 37.7 & 39.6 & 59.2 & 71.6 & 54.1 & 93.0 & 68.2 & 78.3 & 29.3 & 91.5 & 65.0 & 78.2 & 65.8 & 65.1 & 59.5 & 67.7 \\
MPF \cite{alnaggar2021multi} & 55.5 & 31 & 93.4 & 30.2 & 38.3 & 26.1 & 28.5 & 48.1 & 46.1 & 18.1 & 90.6 & 62.3 & 74.5 & 30.6 & 88.5 & 59.7 & 83.5 & 59.7 & 69.2 & 49.7 & 58.1 \\
AMVNet \cite{liong2020amvnet} & 65.3 & - & 96.2 & 59.9 & 54.2 & 48.8 & 45.7 & 71.0 & 65.7 & 11.0 & 90.1 & 71.0 & 75.8 & 32.4 & 92.4 & 69.1 & 85.6 & 71.7 & 69.6 & 62.7 & 67.2 \\

\hline
SPVCNN \cite{tang2020searching} & 63.8 & 187 & - & - & - & - & - & - & - & - & - & - & - & - & - & - & - & - & - & - & - \\
SPVNAS \cite{tang2020searching} & 67.0 & - & 97.2 & 50.6 & 50.4 & 56.6 & 58.0 & 67.4 & 67.1 & 50.3 & 90.2 & 67.6 & 75.4 & 21.8 & 91.6 & 66.9 & 86.1 & 73.4 & 71.0 & 64.3 & 67.3 \\
Cylinder3D \cite{zhu2021cylindrical} & 67.8 & 178 & 97.1 & 67.6 & 64.0 & 59.0 & 58.6 & 73.9 & 67.9 & 36.0 & 91.4 & 65.1 & 75.5 & 32.3 & 91.0 & 66.5 & 85.4 & 71.8 & 68.5 & 62.6 & 65.6 \\
DRINet \cite{ye2021drinet} & 67.5 & 62 & 96.9 & 57.0 & 56.0 & 43.3 & 54.5 & 69.4 & 75.1 & 58.9 & 90.7 & 65.0 & 75.2 & 26.2 & 91.5 & 67.3 & 85.2 & 72.6 & 68.8 & 63.5 & 66.0 \\
AF2S3Net \cite{cheng20212} & 69.7 & - & 94.5 & 65.4 & 86.8 & 39.2 & 41.1 & 80.7 & 80.4 & 74.3 & 91.3 & 68.8 & 72.5 & 53.5 & 87.9 & 63.2 & 70.2 & 68.5 & 53.7 & 61.5 & 71.0 \\
RPVNet \cite{xu2021rpvnet} & 70.3 & $168^*$ & 97.6 & 68.4 & 68.7 & 44.2 & 61.1 & 75.9 & 74.4 & 73.4 & 93.4 & 70.3 & 80.7 & 33.3 & 93.5 & 72.1 & 86.5 & 75.1 & 71.7 & 64.8 & 61.4 \\

\hline
{\bf CPGNet} [ours] & 68.3 & $43 / 35.6^* $ & 96.7 & 62.9 & 61.1 & 56.7 & 55.3 & 72.1 & 73.9 & 27.9 & 92.9 & 68.0 & 78.1 & 24.6 & 92.7 & 71.1 & 84.6 & 72.9 & 70.2 & 64.5 & 71.9 \\

\hline
\end{tabular}
\end{center}
\end{table*}

\begin{table}[t]
\caption{Results on nuScenes validation set.}
\label{table_nuscenes}
\begin{center}
\begin{tabular}{l|c}
\hline
{\bf Methods} & {\bf mIoU} \\

\hline
RangeNet++ \cite{milioto2019rangenetplus} & 65.5 \\
SalsaNext \cite{cortinhal2020salsanext} & 72.2 \\
AMVNet \cite{liong2020amvnet} & 76.1 \\
Cylinder3D \cite{zhu2021cylindrical} & 76.1 \\
RPVNet \cite{xu2021rpvnet} & 77.6 \\

\hline
{\bf CPGNet} [ours] & 76.9 \\
\hline
\end{tabular}
\end{center}
\end{table}

We report results on nuScenes validation set. As shown in Table \ref{table_nuscenes}, CPGNet still outperforms the 2D projection-based methods and can be comparable with the first ranking RPVNet.

\subsection{Ablation Study}
In order to figure out the effectiveness of the proposed components, we make ablation study on SemanticKITTI validation set with the same experimental setup.

First, we make ablative analysis on the Point-Grid (PG) fusion block. As shown in Table \ref{table_ablation1}, the baseline (first row) is the reproduced MPF \cite{alnaggar2021multi}, which adopts our 2D FCN architecture and transformation consistency loss. From the latter rows, we can discover that: 1) the Point Fusion outperforms by +0.9 mIoU compared with the post-process of MPF; 2) it achieves 2.1 gains of mIoU when introducing point feature; 3) the CPGNet with 2 PG fusion blocks leads to the biggest mIoU improvement, compared with the single-block version.

Subsequently, we analyze the effects of other components, including 2D FCN architecture, transformation consistency loss, and TensorRT FP16 deployment, as shown in Table \ref{table_ablation2}. For 2D FCN architecture, dual down-sampling block (DDB) and attention pyramid feature fusion (APFN) improve the mIoU by 0.3, 0.7, respectively, demonstrating their effectiveness. In the experiment, TTA augments point cloud for three times, namely flipping along $x$ axis, flipping along $y$ axis, and flipping along the both. As can be seen, TTA improves the performance (+1.2 mIoU) but needs model inference for four times. When we add transformation consistency loss during training and remove TTA during inference, it performs slightly better (+0.2 mIoU) than TTA with model inference for once. Besides, CPGNet can be easily deployed on TensorRT FP16 inference mode and runs 26.8\,ms per scan with negligible performance drop (-0.1 mIoU).

\begin{table}[t]
\caption{Point-Grid (PG) fusion block analysis on SemanticKITTI validation set.}
\label{table_ablation1}
\setlength\tabcolsep{2.8pt}
\begin{center}
\begin{tabular}{ccc|c|c}
\hline
{\bf Point Fusion} & {\bf Point Feature} & {\bf Blocks} & {\bf mIoU} & {\bf speed (ms)} \\
\hline
$\times$ & $\times$ & 1 & 59.5 & 19.2 \\
\checkmark & $\times$ & 1 & 60.4 & 18.6 \\
\checkmark & \checkmark & 1 & 62.5 & 21.7 \\
\checkmark & \checkmark & 2 & 65.9 & 43 \\

\hline
\end{tabular}
\end{center}
\end{table}

\begin{table}[t]
\caption{Effects of 2D FCN architecture and transformation consistency loss on SemanticKITTI validation set. DDB: Dual down-sampling block. AFPN: Attention feature pyramid fusion. TTA: Test time augmentation. TRT16: TensorRT FP16}
\label{table_ablation2}
\setlength\tabcolsep{2.8pt}
\begin{center}
\begin{tabular}{ccccc|c|c}
\hline
{\bf DDB} & {\bf AFPN} & {\bf TTA} & {\bf $\mathcal{L}_{tc}$} & {\bf TRT16} & {\bf mIoU} & {\bf speed (ms)} \\
\hline
$\times$ & $\times$ & $\times$ & $\times$ & $\times$ & 63.5 & 38.4 \\
\checkmark & $\times$ & $\times$ & $\times$ & $\times$ & 63.8 & 39.2 \\
\checkmark & \checkmark & $\times$ & $\times$ & $\times$ & 64.5 & 43 \\
\checkmark & \checkmark & \checkmark & $\times$ & $\times$ & 65.7 & 172 \\
\checkmark & \checkmark & \checkmark & \checkmark & $\times$ & 66.1 & 172 \\
\checkmark & \checkmark & $\times$ & \checkmark & $\times$ & 65.9 & 43 \\
\checkmark & \checkmark & $\times$ & \checkmark & \checkmark & 65.8 & 26.8 \\

\hline
\end{tabular}
\end{center}
\end{table}

\section{CONCLUSIONS}
In this paper, we present the accurate, fast, and easy-deployed CPGNet for LiDAR semantic segmentation, where point, bird’s-eye view and range view features are fused in a cascade framework. The transformation consistency loss is proposed to save the inference time without performance drop compared with TTA. Besides, we find that 3D point features are beneficial for keeping complete point cloud information, while 2D grid features are suitable for efficient semantic feature extraction.


\bibliographystyle{IEEEtran}
\bibliography{IEEEexample}

\end{document}